\documentclass[ngerman,english]{article}
\usepackage[T1]{fontenc}
\usepackage[latin9]{inputenc}
\usepackage{array}
\usepackage{textcomp}
\usepackage{amsmath}

\makeatletter

\providecommand{\tabularnewline}{\\}

\@ifundefined{date}{}{\date{}}
\makeatother

\usepackage{babel}
\begin{document}
\title{Space of Reasons and Mathematical Model\\
{\normalsize{}Logical Understanding III}}
\author{Florian Richter}
\maketitle
\begin{abstract}
Inferential relations govern our concept use. In order to understand
a concept it has to be located in a ``space of implications'' (Wilfrid
Sellars). There are different kinds of conditions for statements,
i.e. that the conditions represent different kinds of explanations,
e.g. causal or conceptual explanations. The crucial questions is:
How can the conditionality of language use be represented. The conceptual
background of representation in models is discussed and in the end
I propose how implications of propositional logic and conceptual determinations
can be represented in a model of a neural network.
\end{abstract}

\section{Introduction}

\subsection{Context \textendash{} Commitments}

In the last paper it was shown that conceptual relations exist within
a space of implications. This means that statements are embedded in
a logical space of conditions. There are different kinds of conditions
for statements: they can represent reasons, conceptual relations,
causes, or also motivational states, which explain actions. Inferring
from the statements to its conditions is part of abductive reasoning.
Charles Sanders Peirce is very honest about his own mistake of confusing
inductive reasoning and abductive reasoning (hypothesis): ``Only
in almost everything I printed before the beginning of this century
I more or less mixed up Hypothesis and Induction...''\cite{Peirce1958}
(CP 8.227) This mix up is very common. 

\subsection{Problem}

Abductive inferences are also often forgotten in the context of reasoning.
Behind the confusion of induction and abduction lies the problematic
point of mathematical representation of reasoning. Induction is easier
to represent and can be based on statistical tools. Markov logic e.g.
is a combination of statistics and predicate logic. Universal formulas,
e.g. ``All M are K.'' ($\forall x(xM\rightarrow xK)$), can represent
rules or conceptual relations (characteristics). But what happens
if we encounter a counterexample? Only one counterexample makes them
false, but if they are taken as weighted, valid, statistical regularities,
this problem might be bypassed. Therefore, if a (possible) world violates
one formula, then the probability of the formula of being true goes
down. If there are only few worlds violated, the formula (statement)
might probably be true. The worlds are constraints on the universally
quantified formulas: ``A first-order KB can be seen as a set of hard
constraints on the set of possible worlds: if a world violates even
one formula, it has zero probability.'' Domingos et al. (2016) state
that the ``basic idea in Markov logic is to soften these constraints:
when a world violates one formula in the KB it is less probable, but
not impossible. The fewer formulas a world violates, the more probable
it is. Each formula has an associated weight that reflects how strong
a constraint it is: the higher the weight, the greater the difference
in log probability between a world that satisfies the formula and
one that does not, other things being equal.''\cite{Domingos2016}

The crucial question is here, how the similarities or differences
are calculated and with regard to which concept or idea. This is a
question from abductive reasoning. E.g. a plastic tree can be in size
and form similar to a real oak tree and is therefore \emph{with regard
to} size and form more similar than a bonsai tree, but \emph{with
regard to} the material the bonsai tree is more similar to an oak
tree. Abductive reasoning is conceptually richer and can be found
often in the field of engineering and in philosophy of technology.
The following table is from Christoph Hubig\cite{Hubig2006} (208):\\

\noindent %
\begin{tabular}{|>{\raggedright}p{3cm}||>{\raggedright}p{4cm}||>{\raggedright}p{4cm}|}
\hline 
 & Perceptual abduction & Conceptual abduction (middle term)\tabularnewline
\hline 
\hline 
Inference on the case & object of perception & characteristics as relevant for building a class\tabularnewline
\hline 
\hline 
Inference on the rule & rule of perception & class/intension of the concept\tabularnewline
\hline 
\hline 
Inference on the best explanation & explanation of perception & conceptual rule of subsumption\tabularnewline
\hline 
\hline 
Inference on the best strategy of explanation & strategy of perception & system of classification\tabularnewline
\hline 
\end{tabular}\\
\\
\\
\begin{tabular}{|>{\raggedright}p{3cm}||>{\raggedright}p{4cm}||>{\raggedright}p{4cm}|}
\hline 
 & Causal abduction & Presuppositional abduction\tabularnewline
\hline 
\hline 
Inference on the case & cause & means, instrument as reliable/proven\tabularnewline
\hline 
\hline 
Inference on the rule & lawlike connections & techniques as yielding results\tabularnewline
\hline 
\hline 
Inference on the best explanation & theories & sciences, technologies\tabularnewline
\hline 
\hline 
Inference on the best strategy of explanation & paradigms, patterns of interpretation & general principle of relation to the world, conception of technology\tabularnewline
\hline 
\end{tabular}\\
\\

\noindent How can these different kinds of inferences be represented
in mathematical models? One example will be picked out to shape the
kind of mathematical thinking behind causal reasoning.

\section{Causal Reasoning \textendash{} Explanations}

In statistics one tries to identify the confounder not as a condition
that messes up the data. The confounder is therefore not the real
cause. The real cause has to be revealed in order to make the data
``meaningful''.\cite{Pearl2018} According to Judea Pearl, statistics
lacks the idea of causality and statisticians are only interested
in correlation. That is a strong statement that might not be applicable
for every statistician. Pearl writes that ``{[}d{]}ata are the ingredients
that go into the estimand recipe. It is critical to realize that data
are profoundly dumb about causal relationships. They tell us about
quantities like P(L | D) or P(L | D, Z). It is the job of the estimand
to tell us how to bake these statistical quantities into one expression
that, based on the model assumptions, is logically equivalent to the
causal query \textendash{} say, P(L | do(D)).'' According to him
the ``whole notion of estimands does not exist in traditional methods
of statistical analysis. There, the estimand and the query coincide.''
Now, if we e.g. would be ``interested in the proportion of people
among those with Lifespan L who took the Drug D, we simply write this
query as P(D | L). The same quantity would be our estimand.'' This
should require ``no causal knowledge. For this reason, some statisticians
to this day find it extremely hard to understand why some knowledge
lies outside the province of statistics and why data alone cannot
make up for lack of scientific knowledge.''\cite{Pearl2018} (14/15)

The role of this scientific knowledge has to be explained. Pearl thinks
of a ``ladder of causation'' that leads from ``correlation'' (pure
statistics) to ``invervention'', which means to do to something
and therefore to intervene in the chain of events, and the highest
point is represented by ``counterfactuals'' that depict the intervention
and are based on causal models about the world.\cite{Pearl2018} (17)
Statistically, events or data ``only'' correlate, but they need
to be seen through ``causal models'', which are established through
interventions, and serve to obtain causal knowledge, that predicts
effects. This is shall be done by the \textquotedblleft algorithmization
of counterfactuals\textquotedblright .\cite{Pearl2018} (9) 

Pearl talks about structural models that have causal assumptions about
the world. Also, in physics they are using models to represent the
world, but do they use causal models, as a layperson might assume.
Max Planck sees that the physical world view with its determined systems
is a model and as a model it is an idealization, because normally
there are measurement errors that could lead to (slightly) different
results than what the model might predict. Measuring the quantities
that are represented in the equations yields not always ``exact''
results. The physical model of the world is at least not affected
by inexact measurements.\cite{Scheibe2007} \textendash{} Well, the
role of quantum mechanics as an indetermined physical model has to
be left out here.

What role does causality play in physics? And how do we usually use
the term? We seem to know it from our everyday talk, when we say things
like: ``Peter came late, because of the heavy traffic.'' So, the
heavy traffic is the cause. We use causal connections to explain the
behavior of persons or also the behavior of objects. Causality is
a concept that is often used in philosophy, but not really in physics,
even though we would expect that it plays a prominent role there.\cite{Scheibe2007}
We could e.g. say that a force or an event is the cause of a movement
or of another event. Strangely, physicists do not use causality in
that way. There is a difference according to Erhard Scheibe between
the causality of states that are determined clearly and an irreversible
causality of events. David Hume and Immanuel Kant examine the irreversible
form of causality of events, that has at its core the assymetry of
the earlier and later. But already Isaac Newton and then Heinrich
Hertz see the equations as the essence of the necessary link between
states. The system is determined and can therefore be exhibited by
the formation of equations which depict or represent quantities. The
system is made calculable by quantification. If something is messing
up the system \textendash{} a confounder \textendash{} it is seen
as the cause that leads to a certain effect, but the cause is yet
not integrated into the system to represent clearly its determination
by an equation. The determination of the system by an equation is
made clearer, when more an more confounders, which might have an influence
on the system, are integrated in order to sustain its calculability
and thus its predictability or can be disregarded as irrelevant.\cite{Scheibe2007}

Models are internally determined and they are represented by inferential
and differential relations\cite{Kambartel2005}, like the equations
represent determined models. Usually, we use causality as an explanation
for the behavior of persons (we produce something and are the cause
or the producer of the cause (see the intervention part in Pearl))
and we use causality as an explanation in an analog manner to explain
the behavior of objects, like the movements of the planets\cite{Kambartel2005},
but explanations in physics do not use the concept causality. Their
models try to get rid of confounders or defeasors in order to have
a ``consistent'' model.

\section{Limits of Models as Means of Representation}

In the debates in the philosophy of science, according to Hubig, the
modeling of dispositional predicates in if-then-clauses, like the
implication fails, because the implication would be true, if the antecendent
is not realized or not realizable (``under-determination'') or have
to be discarded, if the consequent has not been realized under the
conditions of the antecendet (``over-determination''). Within these
debates the approaches in the philosophy of science change to counterfactual
conditionals as means of representation and search for prognostic
or indicative sentences, which are equivalent to them. To avoid that
the prognostic or indicative sentences turn to be counterintuitive,
pragmatic conditions of the antecedent have to be added, which correspond
to ``conceptions of normality''. In this way, the validity of prognostic
and indicative sentences can be guaranteed or warranted.\foreignlanguage{ngerman}{\cite{Hubig2010}}

Such conceptions of standardized cases cannot be represented by a
complete list of conditions in the form of explicit rules. If that
would be possible, then a set of premises, which would represent a
complete list of the circumstances or conditions ($\varSigma$) as
well as a rule as the major premise ($\Phi$), would always lead to
an instantiation of the rule ($\varphi$) \textendash{} this is a
kind of deductive reasoning ($\varSigma,\Phi\Rightarrow\varphi$\foreignlanguage{ngerman}{).}\footnote{\selectlanguage{ngerman}%
The greek letters represent names in the metalanguage.\selectlanguage{english}%
}\foreignlanguage{ngerman}{\cite{Kambartel2005} (213)}

Hubig points out that we can not describe ``complete states of the
world\foreignlanguage{ngerman}{''\cite{Hubig2010}} (here $\varSigma$),
but it is at least possible to describe a set of ``potential defeasors''\cite{Brandom2008}
(107)\foreignlanguage{ngerman}{}\footnote{\selectlanguage{ngerman}%
It is worth to mention that ``the potential defeasors are \emph{not}
limited to sentences that are \emph{true}'', because a ``non-actual
state of affairs is \emph{possible}''. (125)\selectlanguage{english}%
} \textendash{} also conditions of the antecedent, but the ones that
should be avoided \textendash{} like Robert Brandom writes, who, independently
of Hubig, with reference to his ideas about ``material inference'',
sheds light on this issue. He states that the material inferences
are ``in general \emph{non-monotonic}''.\foreignlanguage{ngerman}{\cite{Brandom2008}
(106)} For him, that does not mean that the ``potential defeasors''
should be made explicit in a complete list. The task is therefore
not to transform such inferences in monotonic ones.\cite{Brandom2008}
(107)\foreignlanguage{ngerman}{}\footnote{\selectlanguage{ngerman}%
``And no one supposes that such probative reasoning can always be
turned into dispositive reasoning by making an explicit, exhaustive
list of the potential defeasors.'' Or: ``There need be no definite
totality of possible defeasors, specifiable in advance. Even where
we have some idea how to enumerate them, unless those provisos are
generally left implicit, actually stating the premises so as to draw
inferences from them monotonically is impossibly cumbersome in practice.''\selectlanguage{english}%
}\foreignlanguage{ngerman}{ ``The potential defeasors in this way
associated with each material inference endorsed in turn define (by
complementation) the range of counterfactual robustness practically
associated with that inference.''\cite{Brandom2008} (108)}

That is why the ``counterfactual robustness'' and the ``potential
defeasors'' are necessary in order to understand the ``conceptual
content of sentences'' or: \foreignlanguage{ngerman}{``counterfactually
robust inferences are an essential aspect of the articulation of the
conceptual contents of sentences''.\cite{Brandom2008} }(125) Brandom
concludes from this, \foreignlanguage{ngerman}{``that in view of
the non-monotonicity of material inference, the practical task of
updating the rest of one\textasciiacute s beliefs when some of them
change is tractable in principle only if those who deploy a vocabulary
practically discriminate ranges of counterfactual robustness for many
of the material inferences they endorse.''\cite{Brandom2008} (109)}

Within such a framework it is not necessary to establish a set of
circumstances or conditions, which (1.) describe a \textquoteleft complete'
state of the world and that is why it is not necessary to allow (2.)
in a formal way an \textquoteleft infinite' list of possible monotonic
premises. One would either take the position of an absolute-deterministic
or a formal and abstract standpoint of an empty metalanguage. From
the first standpoint there would be the requirement, that there are
no possible conditions that could serve as defeasors, which have to
be subtracted or incorporated in order to assure the result of the
inference. This would mean that \emph{all} conditions are explicit.
The last standpoint would allow to add every true premise like in
a formal and monotonic logic. \textendash{} Another way to assure
the result of the consquence, the application or instantiation of
the rule, would be to introduce a meta-rule, which settles the application,
but that would lead to an infinite regress. 

\section{Models \textendash{} Representational Tools}

Theories use models in order to represent structures. The models encompass
an object-language that is normally weaker in its expressive power
than the metalanguage\footnote{It might be also the other way around: e.g. an ``extensional metalanguage
for intensional languages'' (``possible world semantics for modality'').\cite{Brandom2008}
(11)}, because the metalanguage is would here be the natural language,
in which we talk about the models and their structures. The models
are not able to do that, because they have no representation of themselves. 

Hubig and Michael Weingarten introduce an important distinction in
the philosophy of science.\cite{Hubig2002,Weingarten2003} Is something
a model \emph{of} or a model \emph{for} something? The ``models of...''
are external ``realizations, instantiations, exemplifications''.
The structure (S) of the realization (R) is revealed via induction
or abduction. If R fulfills every rule of the formulas (F) of S. At
least, this is possible in a axiomatic system.\cite{Hubig2002} (35)
The ``models for...'' are ``paradigmatic abstractions, i.e. one-sided
pictures of structures''. Hubig calls them ``conceptualized models''
or a ``model-idealism''. The only thing that counts is the ``correctness
of the use of the symbols''.\cite{Hubig2006} 

For example, a camera can be a model \emph{of} the eye. The camera
is a technical device that allows to explain the function of the eye.
There is a symmetry assumed between the technical model (device) and
the realization or instantiation of the model (eye). The model and
the representation in the model (eye) belong as instantiations to
the same class of natural laws. The function of both (camera and eye)
falls under the class of the laws of optics.\cite{Weingarten2003}
\textendash{} It is often not seen anymore that the technical device
is used as a model, because we are so accustomed to their use in our
daily life, but also in the sciences.\cite{Weingarten2003}

The presupposed symmetry allows to establish an equivalence, but this
is done on the level of the metalanguage, but for that it has to be
clear that we are using a model to represent something else as something,
e.g. the mind as a computer or a logical machine. Computation and
computability are also certain aspects of human reasoning and were
developed as a model \emph{of} human reasoning. \textendash{} Neural
processes construct the reality and semantic significance. The starting
point would be the neural organization of us and the emergence of
mind and meaning out of it, but the supposed relation of equivalence
is empty as long as the differences are not reflected. This happens
when e.g. a technical device is used to form a model of something
else without reflecting that it is a model, i.e. a tool to represent
something. If we use the model for a certain purpose, the assumed
symmetry has to prove itself in the world/reality. The model is then
the means to accomplish the purpose and has to yield a certain effect.\cite{Hubig2006}
In neural networks the neurons are adjusted through backpropagation
in order to bring the ouput closer to the desired output. One model
for machine learning is the brain and it is therefore the means that
is slotted in between the output and the desired ouput. It is a real
possibility \textendash{} the \emph{medium}, in which the adjustments
are done \textendash{} and not just a formal possibility of manipulating
symbols, because the specific adjustments are represented by the ouput.
It is a representation via the model.

\section{Representations of the Mind: Neural Networks}

\subsection{The Idea of Neural Networks}

The idea is not to rebuild the human brain, but to use a model of
it for machine learning. How are the neurons connected and how do
they make connections? Neural networks are one-sided, paradigmatic
pictures of the functioning of the brain. They are of course models
\emph{of} the brain, but are taken here as models \emph{for} machine
learning. Every neuron (approximately 100 billion in one human brain)
has an axon and is via dendrites connected with other neurons. If
a certain voltage exceeds a threshold the neuron fires.\cite{Ertel2017}
(9.1) How the brain works and how neurons work is still a big task
for researchers with many unsolved riddles. The model is, as said
before, just a paradigmatic picture. 

Warren McCulloch and Walter Pitts proposed in 1943 a model of a neuron.\cite{McCulloch1943}
They propose an equivalence between the functioning of the neurons
of the brain and propositional logic (which can be also represented
by logic gates in computers). McCulloch and Pitts claim that ``neither
of us conceives the formal equivalence to be a factual explanation''.
The physical behaviors of the neurons ``in no way affect the conclusions
which follow from the formal treatment of the activity of nervous
nets'', i.e. the behavior of the neurons does not affect the formal
connections of propositional logic represented by the logical operators.
Propositional logic is the normative structure, which underlies the
physical activity.\cite{McCulloch1943} (101) The determination of
the states of the nervous net is based on necessary connections, which
are irreciprocal (causality\footnote{Causality is here conceived similar to the notion of determination
(which is described above), because a certain kind of consistency
of the system is required.}). There cannot be a complete knowledge (or determination), because
of the incompleteness as to space and indefiniteness as to time, and
the ``inclusion of disjunctive relations prevents complete determination''
on the logical side.\cite{McCulloch1943} (113) Such a causal determination
is equivalent to the logic of propositions and disjunctions can be
added to determine the system step by step. \textendash{} Even though
they are concerned with the logical determination, the notion of disjunctive
relations is based on the calculus of propositional logic and something
else than conceptual determination (see below).

\subsection{Perceptron}

Frank Rosenblatt introduced then in 1958 the idea of a perceptron.
The model is formulated ``in terms of probability theory rather than
symbolic logic.'' The activations of the neurons are calculated via
the ``algebraic sum of excitatory and inhibitory impulse intensities'',
which are ``equal to or greater than the threshold ($\theta$)''.\cite{Rosenblatt1958}
(387-389) Perceptrons faced also several problems and in ``1969,
Minsky and his colleague Seymour Papert published \emph{Perceptrons},
a book detailing the shortcomings of the eponymous algorithm, with
example after example of simple things it couldn\textasciiacute t
learn.'' \cite{Domingos2018} (100) A learning algorithm is judged
by its ability to learn and if there are basic things, that it can
not do, it needs to be disregarded or improved, but the question is
not only about learning, also about the ability to represent things.
The problematic issue, that Pedro Domingos points out, is more about
the expressive richness of the representational model. And in the
case of the perceptron it cannot represent the ``exclusive-OR function''\cite{Domingos2018}
(100), but a mulitlayer network can do it.

\subsection{Hopfield-Model}

John Hopfield, a physicist, suggested in 1982 an equivalence between
neural nets and a physical system of magnetism (spin glasses). Hopfield
writes: 
\begin{quote}
``There are classes of physical systems whose spontaneous behavior
can be used as a form of general (and error-correcting) content-addressable
memory. Consider the time evolution of a physical system that can
be described by a set of general coordinates. A point in state space
then represents the instantaneous condition of the system. This state
space may be either continuous or discrete (as in the case of \emph{N}
Ising spins).

The equations of motion of the system describe a flow in state space.
Various classes of flow patterns are possible, but the systems of
use for memory particularly include those that flow toward locally
stable points from anywhere within regions around those points. A
particle with frictional damping moving in a potential well with two
minima exemplifies such a dynamics.''\cite{Hopfield1982} (2554) 
\end{quote}
Wolfgang Ertel states that in the cases of pattern recognition, which
are learned by the Hopfield-model and are trained by certain examples
with a certain amount of neurons, the states are finite. Like in the
physical system the energy function will reach a minimum, but if there
are to many patterns learned, it can lead to a chaotic dynamic and
the system or neural network is not able to recognize the patterns.
In the model of Hopfield the neurons can learn to recognize patterns,
but only if there are enough neurons in the model, otherwise it changes
from an ordered dynamic to a chaotic dynamic.\cite{Ertel2017} (9.2)
Another problematic point is that the states in the model of Hopfield
are binary. He uses a step function\footnote{``A study of emergent collective effects and spontaneous computation
must necessarily focus on the nonlinearity of the input-output relationship.
The essence of computation is nonlinear logical operations. The particle
interactions that produce true collective effects in particle dynamics
come from a nonlinear dependence of forces on positions of the particles.
{[}...{]} Those neurons whose operation is dominantly linear merely
provide a pathway of communication between nonlinear neurons. Thus,
we consider a network of \textquoteleft on or off' neurons, granting
that some of the interconnections may be by way of neurons operating
in the linear regime.''\cite{Hopfield1982} (2555)}, while the activation of the neuron is now mostly calulated via the
sigmoid function to propagate through the neural network. 

\subsection{Forward Propagation and Backpropagation}

Backpropagating is a learning procedure for neural networks.\cite{Rumelhart1986}
It means to go back layer by layer and to adjust the weights, but
first you need to propagate through the neural network. The calculation
of the output or hypothesis is the sum of the weights:
\begin{description}
\item [{$h_{\theta}(x)=\sigma\left(\sum_{i=1}^{n}\theta_{ji}x_{i}\right)$}]~
\end{description}
$\theta_{ji}$ is the weight or parameter that connects a neuron of
the layer i to a neuron on another layer j. The activation of the
neuron is calculated via the sigmoid function:
\begin{description}
\item [{$\sigma(x)=\frac{1}{1+e^{-x}}$}]~
\end{description}
And if there is now a difference between the desired output and the
real output, a learning algorithm propagates back. The loss function
(sometimes also called cost function) is used to adjust the weights
to minimize the difference between the actual output vector and the
desired output vector.\cite{Rumelhart1986} (533)\footnote{For the backpropagation algorithm see Rumelhart et al. (1986)\cite{Rumelhart1986}
or a textbook like Ertel (2017).\cite{Ertel2017} (9.3.1)}

\section{Neural Networks: Intuition}

The goal is to build a neural network that is able to represent the
logical implication of propositional logic. The compatibilities and
incompatibilities can be expressed in the following way.
\begin{description}
\item [{$p\rightarrow q$}] is incompatible with $\diamondsuit(p\land\lnot q)$
\item [{$p\rightarrow q$}] is compatible with $\diamondsuit(\lnot p\lor q)$
\end{description}
It is therefore possible to use the negation and the disjunction to
build a neural network that has as its output the same truth values
like the implication in propositional logic:\\

\begin{tabular}{|c|c|c|}
\hline 
$p$ & $q$ & $p\rightarrow q$\tabularnewline
\hline 
\hline 
t & t & t\tabularnewline
\hline 
t & f & f\tabularnewline
\hline 
f & t & t\tabularnewline
\hline 
f & f & t\tabularnewline
\hline 
\end{tabular}\\
\\
The inputs are $x_{1}$ and $x_{2}$ (above $p$ and $q$). The second
layer contains $a_{1}^{[2]}$ and $a_{2}^{[2]}$. And $h_{\Theta}(x)$
is the output of the function, which is computed by the variables
and the parameters or weights ($\theta_{1},\theta_{2},...$; as a
matrix: $\Theta$). For this activation function the sigmoid function
is used: $\sigma(x)=\frac{1}{1+e^{-x}}$ and a bias unit has to be
added $b$. The function is thus:
\begin{description}
\item [{$h_{\theta}(x)=\sigma(b+\theta_{1}x_{1}+\theta_{2}x_{2})$}]~
\end{description}
To compute the negation:
\begin{description}
\item [{$h_{\theta}(x)=\sigma(5-10x_{1}+0x_{2})$}]~
\end{description}
To compute the second layer $a_{2}^{[2]}$, which should have the
same value as $x_{2}$:
\begin{description}
\item [{$h_{\theta}(x)=\sigma(-10+0x_{1}+20x_{2})$}]~
\end{description}
To compute the disjunction of the second layer neurons:
\begin{description}
\item [{$h_{\theta}(a)=\sigma(-5+10a_{1}^{[2]}+10a_{2}^{[2]})$}]~
\end{description}
The truth values that are calculated with the sigmoid function are
approximations to 1 and 0. The calculation of the activation of the
neurons has the same truth values like the implication in propositional
logic:

\begin{tabular}{|c|c|c|c|c|}
\hline 
$x_{1}$ & $x_{2}$ & $a_{1}^{[2]}$ & $a_{2}^{[2[}$ & $h_{\theta}(x)$\tabularnewline
\hline 
\hline 
1 & 1 & 0 & 1 & 1\tabularnewline
\hline 
1 & 0 & 0 & 0 & 0\tabularnewline
\hline 
0 & 1 & 1 & 1 & 1\tabularnewline
\hline 
0 & 0 & 1 & 0 & 1\tabularnewline
\hline 
\end{tabular}\\
\\
Incompatibility of an implication ($\diamondsuit(p\land\lnot q)$)
can be represented in a similar way. The neural networks represent
possible connections or states of conceptual relations and if the
input or the information changes the parameters can be adjusted.

It was mentioned that McCulloch and Pitts introduce the notion of
the ``inclusion of disjunctive relations'', which ``prevents complete
determination'' of the states of the neurophysiological and logical
model.\cite{McCulloch1943} (113) I propose the introduction of conjunctions
that should be included in order to have a conceptual determination.
The conjunctions can be added as inputs and connected with the second
layer ($a_{2}^{[2]}$). They are possible determinations:
\begin{description}
\item [{$\lnot p\lor\diamondsuit(q\land r)$}]~
\end{description}
Or:
\begin{description}
\item [{$p\rightarrow\diamondsuit(q\wedge\lnot r)$}]~
\end{description}
These conceptual relations remain, even if $p$ is not the case. The
proposed model can represent implications and determines conceptually. 

\bibliographystyle{plain}
\bibliography{BibliographyLogicalUnderstanding}

\end{document}